\documentclass[conference]{IEEEtran}
\IEEEoverridecommandlockouts
\usepackage{cite}
\usepackage{amsmath,amssymb,amsfonts}
\usepackage{algorithmic}
\usepackage{graphicx}
\usepackage{textcomp}
\usepackage{xcolor}
\usepackage[ruled,linesnumbered]{algorithm2e}
\usepackage{multirow}
\usepackage{enumerate}
\usepackage{url}
\usepackage{booktabs}
\usepackage{threeparttable}
\usepackage{endnotes}
\usepackage{bbding}
\usepackage{bm}
\usepackage{CJKutf8}
\usepackage[utf8]{inputenc}
\usepackage{subfigure}

\def\BibTeX{{\rm B\kern-.05em{\sc i\kern-.025em b}\kern-.08em
    T\kern-.1667em\lower.7ex\hbox{E}\kern-.125emX}}
\begin{document}

\begin{CJK*}{UTF8}{gbsn}

\title{An Attention-based Bi-GRU-CapsNet Model for Hypernymy Detection between Compound Entities}
\author{\IEEEauthorblockN{Qi Wang$^1$, Chenming Xu$^2$, Yangming Zhou$^{1,*}$, Tong Ruan$^{1}$, Daqi Gao$^1$ and Ping He$^{3,*}$}
\IEEEauthorblockA{$^1$School of Information Science and Engineering, East China University of Science and Technology, Shanghai 200237, China\\
$^2$School of Science, East China University of Science and Technology, Shanghai 200237, China\\
$^3$Shanghai Hospital Development Center, Shanghai 200040, China \\
$^*$Corresponding authors\\
Emails:\{ymzhou@ecust.edu.cn, heping@shdc.org.cn\}}
}

\maketitle

\begin{abstract}

Named entities are usually composable and extensible. Typical examples are names of symptoms and diseases in medical areas. To distinguish these entities from general entities, we name them \textit{compound entities}. In this paper, we present an attention-based Bi-GRU-CapsNet model to detect hypernymy relationship between compound entities. Our model consists of several important components. To avoid the out-of-vocabulary problem, English words or Chinese characters in compound entities are fed into the bidirectional gated recurrent units. An attention mechanism is designed to focus on the differences between two compound entities. Since there are some different cases in hypernymy relationship between compound entities, capsule network is finally employed to decide whether the hypernymy relationship exists or not. Experimental results demonstrate the advantages of our model over the state-of-the-art methods both on English and Chinese corpora of symptom and disease pairs.

\end{abstract}


\section{Introduction}
\label{Sec:Introduction}

Hypernymy relationship plays a critical role in language understanding because it enables generalization, which lies at the core of human cognition. Hypernymy detection is useful for natural language processing tasks such as taxonomy creation \cite{Snow2006Semantic}, question answering \cite{McNamee2008} and sentence similarity estimation \cite{Sogancioglu2017}. However, existing methods for hypernymy detection deal with the case where an entity only includes a word. In fact, named entities which composed of multiple continuous words frequently occur in domain-specific areas. These entities are usually composable and extensible. Typical examples are names of symptoms and diseases in medical areas. Take ``\textit{pain}'' as an example, body parts can be added to it, such as ``\textit{head pain}''. Severe words can also be added to it, such as ``\textit{severe head pain}''. Furthermore, causes can be added to it, such as ``\textit{severe head pain due to drug}''. Note that not all the multiple-word entities have this composite property. To distinguish them from general entities, we define them as \textit{compound entities}.

\begin{figure}[t]
\begin{center}
\subfigure[]{
\label{fig:a} 
\includegraphics[width=1.6in]{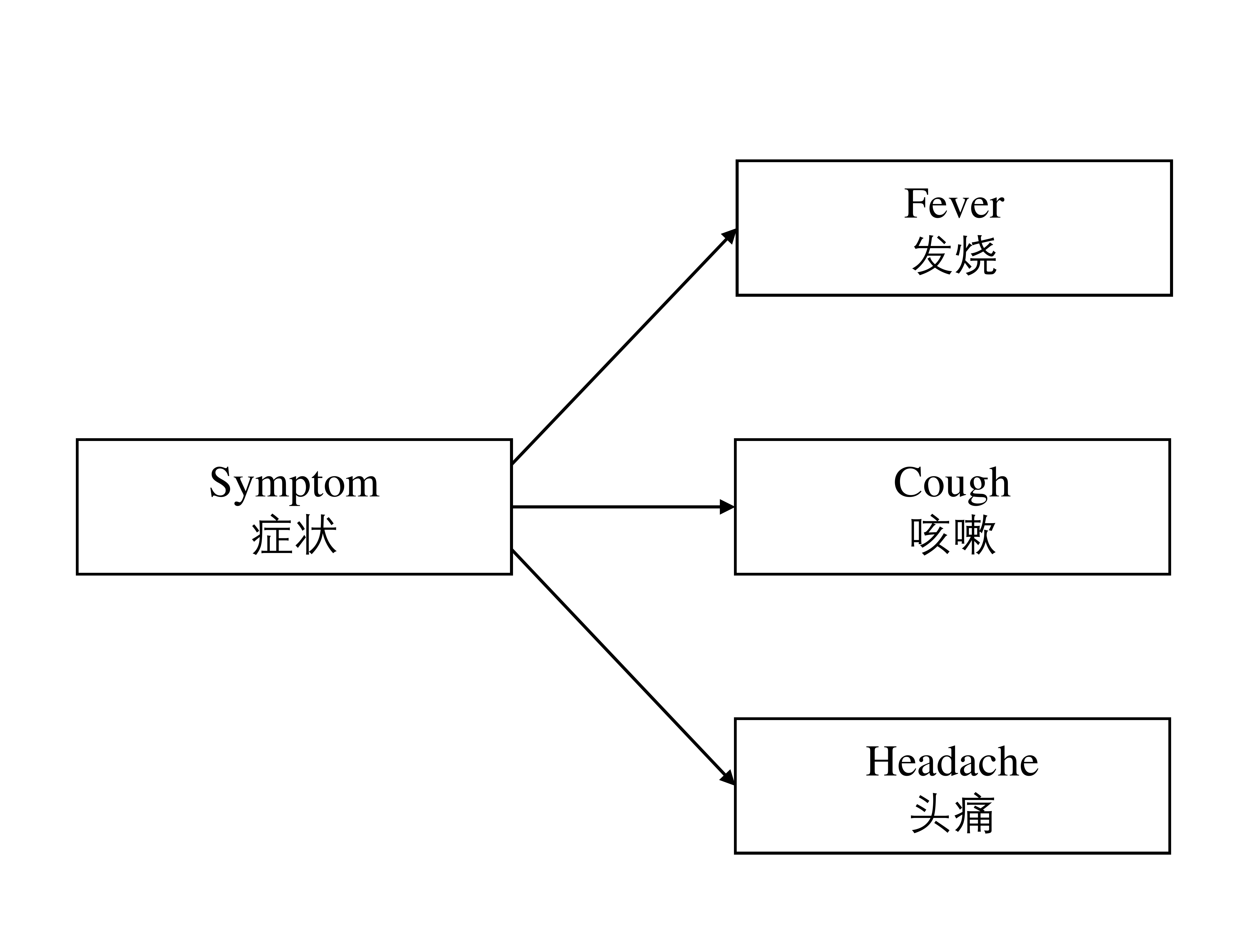}}
\hspace{1.8pt}
\subfigure[]{
\label{fig:b} 
\includegraphics[width=1.6in]{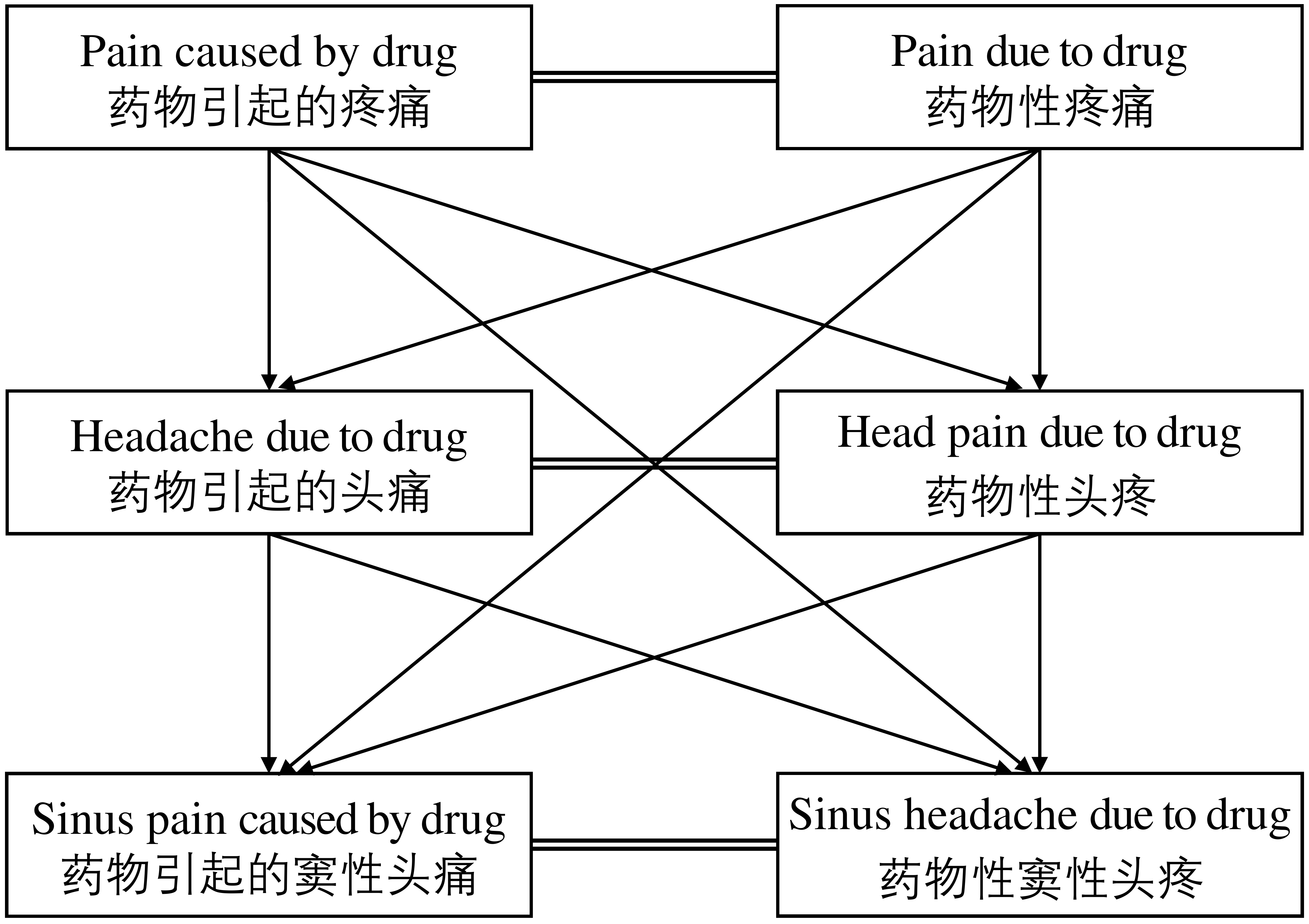}}
\caption{Two examples to respectively explain the hypernymy relations between general entities (see (a) at the leftmost) and compound entities (see (b) at the rightmost). The arrows point from the hypernym to the hyponym and the equality signs indicate the two compound entities are synonymous.}
\label{Fig:Hypernymy Relation Examples}
\end{center}
\end{figure}

Given a pair of entities $(X_1 ,X_2)$, hypernymy detection aims to check if $X_1$ names a broad category that includes $X_2$. If this relation holds between entities $X_1$ and $X_2$, we call $X_1$ is a hypernymy of $X_2$ and $X_2$ is a hyponym of $X_1$. Unlike the hypernymy relationship between two general entities, this work aims to detect hypernymy relations between compound entities. To illustrate it, we provide examples in Fig. \ref{Fig:Hypernymy Relation Examples}.

Hypernymy detection is a very challenging research topic. A number of methods have been proposed for hypernymy detection. These methods are usually based on lexico-syntactic paths \cite{hearst1992automatic,snow2005learning,shwartz2016improving}, distributional representations \cite{kotlerman2010directional,lenci2012identifying,Fu2013Exploiting}, or distributed representations \cite{yu2015learning,Anh2016Learning}. However, these methods are proposed for general entities. While for the hypernymy detection between compound entities, existing methods have two main deficiencies. On the one hand, these methods suffer from the Out-Of-Vocabulary (OOV) problem (that compound entities may not appear in the training set). On the other hand, these methods do not take into account different cases in the hypernymy relationship between compound entities \cite{wang2018using}.

In this paper, we propose an attention-based Bi-GRU-CapsNet model to detect hypernymy relationship between compound entities. Our model integrates several important components. Firstly, English words or Chinese characters in compound entities are fed into Bidirectional Gated Recurrent Units (Bi-GRUs). Secondly, an attention mechanism is used to focus on the differences between two compound entities. Finally, Capsule Network (CapsNet) is employed to decide whether there is a hypernymy relationship between the entity pair. We assess the performance of the proposed model on both Chinese and English corpora of symptom and disease pairs.
The main contributions of this work can be summarized as follows:
\begin{itemize}
    \item We define the concept of compound entities, and propose an attention-based Bi-GRU-CapsNet model to detect hypernymy relations between compound entities. The model is based on the internal elements of compound entities, which does not require the contextual information.
    \item We build two English and Chinese corpora of symptom and disease pairs for the hypernymy detection between compound entities.
    \item Experimental results demonstrate that our attention-based Bi-GRU-CapsNet model outperforms the state-of-the-art methods both on Chinese and English corpora of symptom and disease pairs.
\end{itemize}

\section{Related work}
\label{Sec:Related Work}

Hypernymy detection is a long-standing research topic. We briefly review three major approaches, namely path-based methods, distributional methods and distributed methods.

Path-based methods identify hypernymy relationship through the lexico-syntactic paths which connect the joint occurrences of entity pairs in a large corpus. The pioneer work is proposed by Hearst \cite{hearst1992automatic}, who has found out that linking two noun phrases via certain lexical constructions often implies hypernymy relationship. Variations of pattern-based methods are later proposed \cite{snow2005learning}. Recently, deep learning methods are also employed \cite{shwartz2016improving}.
Path-based methods are simple and efficient. However, due to ambiguity of natural language, it is not robust to detect the hypernymy relationship according to the context of entity pairs. Furthermore, people do not express every possible hypernymy relationship in natural-language texts. It limits the recall of these methods. The path-based methods require co-occurrence of an entity pair, but there are a few hypernymy pairs of compound entities in the same sentence. Thus, path-based methods are not applicable to hypernymy detection between compound entities.

Distributional methods represent a category of methods which use the distributional representations of entity pairs, i.e. the contexts with which each entity occurs separately in the corpus. Some studies follow the distributional inclusion hypothesis \cite{kotlerman2010directional,lenci2012identifying}. Some studies assume the hypernyms of an entity co-occur with it frequently \cite{Fu2013Exploiting}. Recently, with the popularity of word embeddings, most focus has shifted towards supervised distributional methods \cite{roller2014inclusive,fu2014learning,Glava2017Dual}.
In contrast with path-based methods, distributional methods do not require co-occurrence of an entity pair, and these methods usually perform better than path-based methods. However, many times a compound entity is more like a sentence rather than a word. When representing compound entities with distributional vectors through their contextual information, these methods often suffer from the Out-Of-Vocabulary (OOV) problem because of the absence of compound entities in the corpus.

Distributed methods utilize neural models to learn distributed representations via pre-extracted hypernymy pairs. Yu et al. \cite{yu2015learning} proposed a supervised method with negative sampling for hypernymy identification. Tuan et al. \cite{Anh2016Learning} further proposed a dynamic weighting neural model to learn term embeddings.
Compared to distributional methods which obtain entity vectors through their contextual information, distributed methods can train entity vectors only by pre-extracted hypernymy pairs. However, previous distributed methods only designed to deal with single-word entities, do not take compound entities into account. Our proposed attention-based Bi-GRU-CapsNet model (see Section \ref{Sec:Attention-Based Bi-GRU-CapsNet}) is also a distributed method. It considers the internal elements of compound entities. English words or Chinese characters in compound entities are fed into Bi-GRUs in order to avoid the OOV problem.

\section{Attention-based Bi-GRU-CapsNet model}
\label{Sec:Attention-Based Bi-GRU-CapsNet}

\begin{figure}[!htb]
\centering
\includegraphics[width=2.97in]{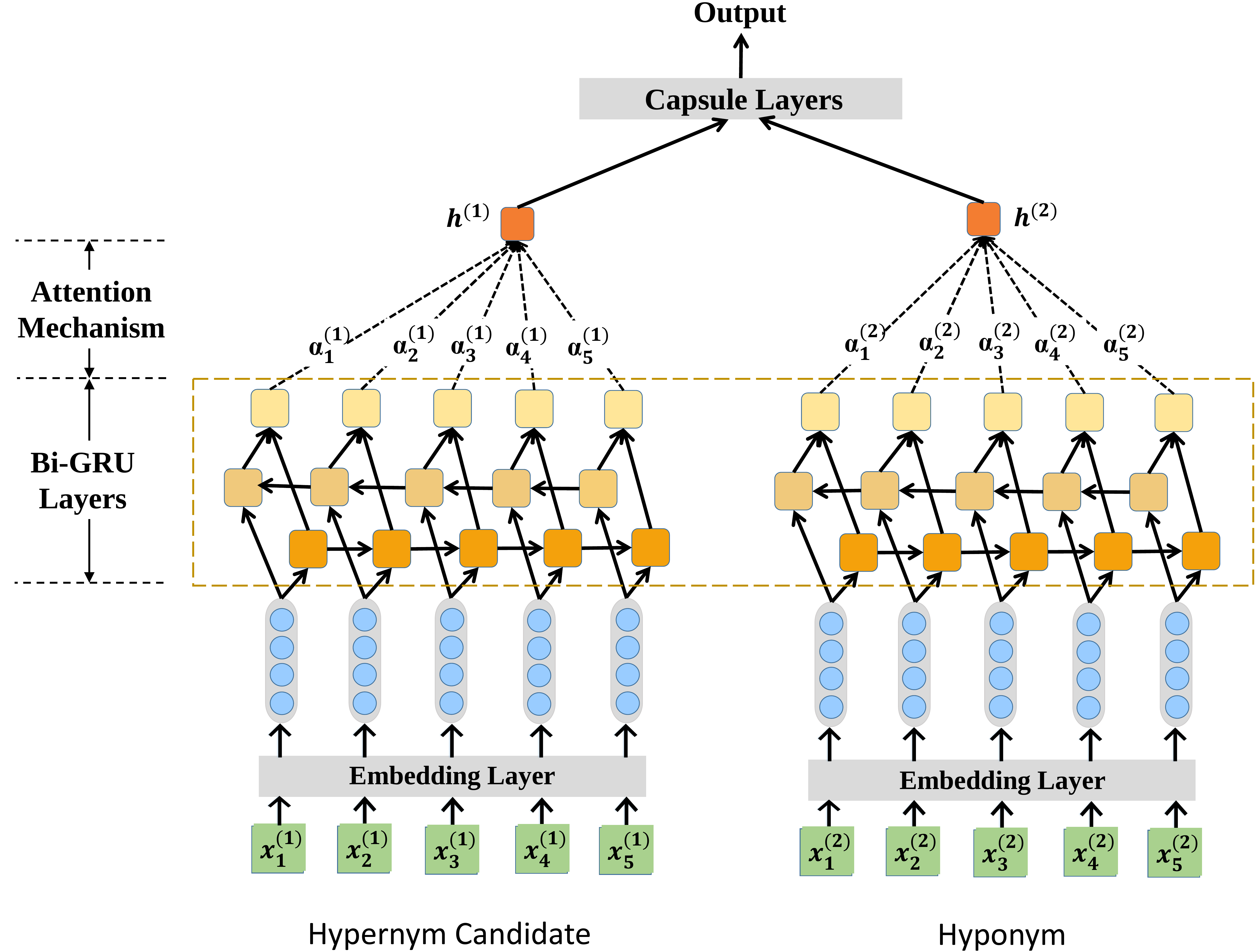}
\caption{The architecture of the attention-based Bi-GRU-CapsNet model.}
\label{fig:model}
\end{figure}

In this section, we present an attention-based Bi-GRU-CapsNet model for hypernymy detection between compound entities. In our model, English words or Chinese characters are first represented as distributed embedding vectors through embedding layers, and then they are fed into the Bi-GRU layers. Afterwards, an attention mechanism is utilized to obtain the feature vector (also considered as a capsule) of each entity. Finally, Capsule network (CapsNet) is used to decide whether the two compound entities have hypernymy relationship. The architecture of our model is shown in Fig. \ref{fig:model}.

\subsection{Embedding layer}

Given a hypernym candidate $X_1$, which is a sequence of $T_1$ words, and a hyponym $X_2$, which is a sequence of $T_2$ words\footnote{$X_1$ and $X_2$ can be sequences of English words or Chinese characters. We say ``words'' in this paper for convenience.}, the first step is to map discrete language symbols to distributed embedding vectors. Formally, we lookup embedding vector from embedding matrix for each word $x_t^{(k)}$ as $\bm{e}_t^{(k)} \in \mathbb{R}^{d_e}$, where $k \in \{1,2\}$ indicates whether the compound entity is a hypernym candidate or a hyponym, $t \in \{1,2,\ldots,T_k\}$ indicates $x_t^{(k)}$ is the $t$-th word in $X_k$, and ${d_e}$ is a hyper-parameter indicating the size of word embedding.

\subsection{Bi-GRU layer}

As a variant of the standard recurrent neural network (RNN), the gated recurrent unit (GRU) was originally proposed by Cho et al. \cite{cho2014learning}. For each position $t$, GRU computes $\bm{h}_t$ with input $\bm{x}_t$ and previous state $\bm{h}_{t-1}$, as:
\begin{equation}
\bm{r}_t=\sigma(\bm{W}_r\bm{x}_t+\bm{U}_r\bm{h}_{t-1})
\end{equation}
\begin{equation}
\bm{u}_t=\sigma(\bm{W}_u\bm{x}_t+\bm{U}_u\bm{h}_{t-1})
\end{equation}
\begin{equation}
\tilde{\bm{h}_t}=\tanh(\bm{W}_c\bm{x}_t+\bm{U}(\bm{r}_t\odot \bm{h}{t-1}))
\end{equation}
\begin{equation}
\bm{h}_t=(1-\bm{u}_t)\odot \bm{h}_{t-1}+\bm{u}_t\odot\tilde{\bm{h}_t}
\end{equation}
where $\bm{h}_t$, $\bm{r}_t$ and $\bm{u}_t$ are $d$-dimensional hidden state, reset gate, and update gate, respectively; $\bm{W}_r$, $\bm{W}_u$, $\bm{W}_c$ and $\bm{U}_r$, $\bm{U}_u$, $\bm{U}$ are the parameters of the GRU; $\sigma$ is the sigmoid function, and $\odot$ denotes element-wise production.

For a word at $t$, we use the hidden state $\overrightarrow{\bm{h}_t}$ from the forward GRU as a representation of the preceding
context, and the $\overleftarrow{\bm{h}_t}$ from the backward GRU that encodes
text reversely, to incorporate the context after $t$. We use the concatenation $\bm{h}_t=[\overrightarrow{\bm{h}_t}; \overleftarrow{\bm{h}_t}]$, the bi-directional contextual encoding of $\bm{x}_t$, as the output of the Bi-GRU layer at $t$.

\subsection{Attention mechanism}

As to hypernymy detection, it is helpful to focus on only the different parts between two compound entities. In our model, we introduce an attention mechanism to improve the detection performance. The feature vector $\bm{h}^{(k)}$ of the entity $X_k$ is defined as a weighted sum, which is computed as follows:
\begin{equation}\label{eq:attention}
\bm{h}^{(k)}=\sum_{i=1}^{T_k}\alpha_i^{(k)} \bm{h}_i^{(k)}
\end{equation}
where $\bm{h}_i^{(k)}$is the output of the Bi-GRU layer at $i$, and the weight $\alpha_i^{(k)}$ is computed by

\begin{equation}
\alpha_i^{(k)}=aw_i^{(k)}+b
\end{equation}

\begin{equation}
w_i^{(k)}=
\begin{cases}
0,&  x_i^{(k)}\in seqLCS\\
1,&  x_i^{(k)}\notin seqLCS
\end{cases}
\end{equation}
Here, $seqLCS$ is the longest common subsequence of $X_1$ and $X_2$, $x_i^{(k)}$ is the $i$-th word of $X_k$, and $a,b$ are the parameters of the network.

As mentioned above, the key to hypernymy detection is to focus on the differences between two compound entities. Thus, it is a better choice that different words between entity pairs contribute more to the final entity vectors $\bm{h}^{(k)}$. Since $\bm{h}^{(k)}$ is a weighted sum (see Eq. \ref{eq:attention}), the attention mechanism aims that the different parts of entity pairs have larger attention weights compared with the overlapping parts. It is like paying more attention to the different words and ignoring the overlapping words, so we call it attention mechanism. Through the attention mechanism, the model can also deal with unseen entity pairs if the different parts have appeared in the training set.

\subsection{Capsule layer}

The capsule layer was originally proposed in \cite{Sabour2017Dynamic} for digit recognition, in which a capsule is a group of neurons whose activity vector represents the instantiation parameters of a specific type of entity. The length of the activity vector is used to represent the probability that the entity exists and its orientation to represent the instantiation parameters. Thus, a non-linear squashing function is used to ensure that short vectors get shrunk to almost zero length and long vectors get shrunk to a length slightly below 1:
\begin{equation}
\bm{v}_j = \frac{{||\bm{s}_j||}^2}{1+{||\bm{s}_j||}^2} \frac{\bm{s}_j}{||\bm{s}_j||}
\end{equation}
where $\bm{v}_j$ is the vector output of input capsule $\bm{s}_j$.

\begin{figure}[!ht]
\centering
\includegraphics[width=2.1in]{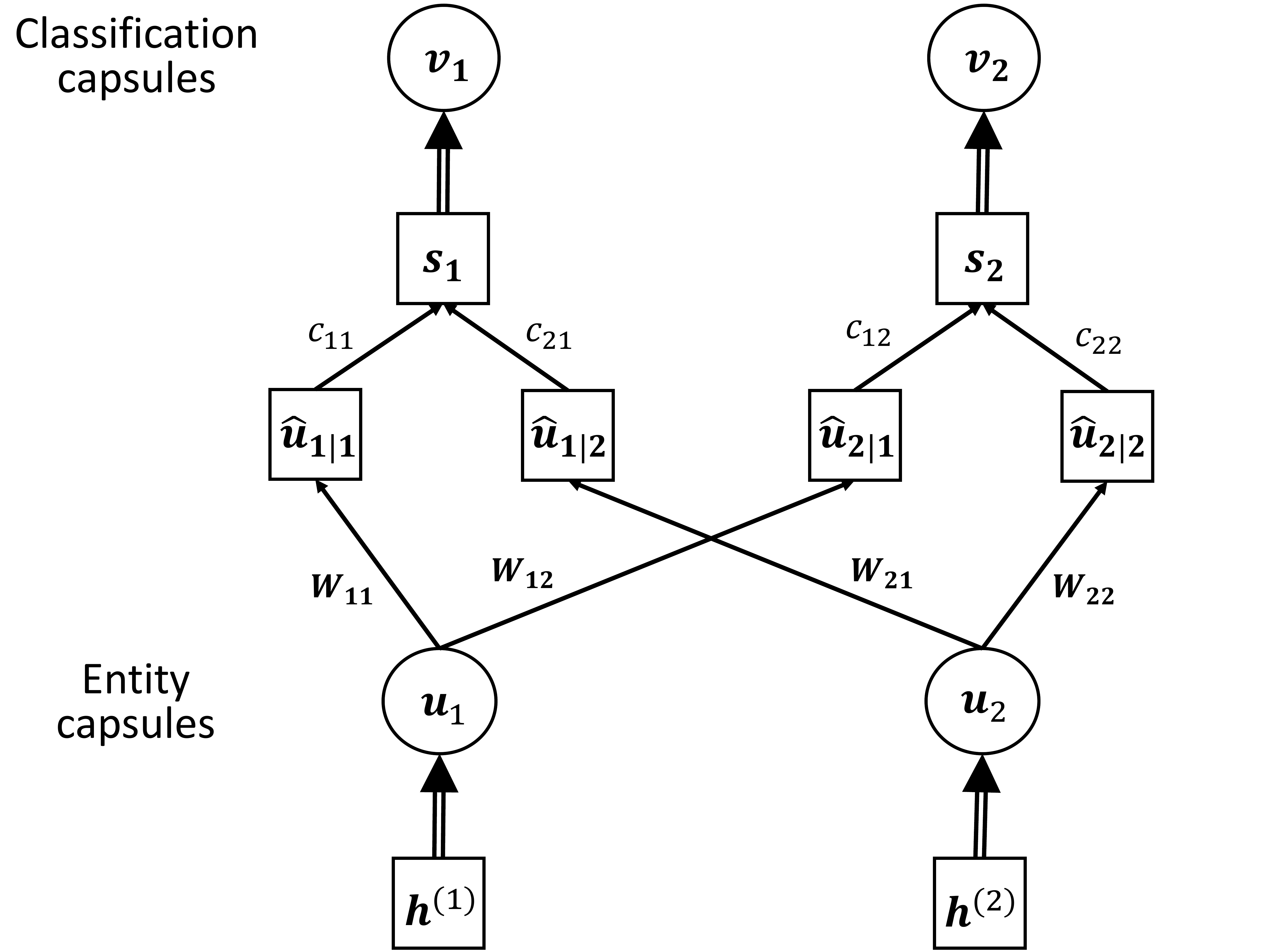}
\caption{Capsule layers (the double-line arrows indicate squashing functions).}
\label{fig:capsLayers}
\end{figure}

Unlike \cite{Sabour2017Dynamic}, we use capsules to represent both entities and relations. Therefore, the probability of hypernymy relationship can be represented by the length of corresponding capsules, and different cases of hypernymy relationship can be represented by the orientation of the capsule. As illustrated in Fig. \ref{fig:capsLayers}, the total input to a capsule $\bm{s}_j$ is a weighted sum over all ``prediction vectors'' $\bm{\widehat{u}}_{j|i}$ from the capsules in the layer below and is produced by multiplying the output $\bm{u}_i$ of a capsule in the layer below by a weight matrix $\bm{W}_{ij}$:

\begin{equation}
\bm{s}_j=\sum_{i=1} c_{ij}\bm{\widehat{u}}_{j|i}
\end{equation}

\begin{equation}
\bm{\widehat{u}}_{j|i} = \bm{W}_{ij} \bm{u}_i
\end{equation}
where $c_{ij}$ are coupling coefficients that are determined by an iterative dynamic routing algorithm with a given number of iterations $r$ (see \cite{Sabour2017Dynamic} for more details).

In training phase, we minimize a separate margin loss $L_j$ for each classification capsule $\bm{v}_j$:
\begin{equation}
L_j = R_j~{\max(0, m^{+}-||\bm{v}_j||)}^2+(1-R_j)~{\max(0, ||\bm{v}_j||-m^{-})}^2
\end{equation}
where $j =1$ means there is a hypernymy relationship between two compound entities, otherwise $j = 2$. $R_j = 1$ iff the corresponding relation $j$ exists and $m^{+} = 0.9$ and $m^{-} = 0.1$. The total loss is simply the sum of the losses of both classification capsules.

\section{Computational results}
\label{Sec:Computational Results}

\subsection{Datasets}
\label{SubSec:Datasets}

Our proposed model is evaluated on English and Chinese corpora\footnote{They are publicly available at \url{https://github.com/ECUST-NLP-Lab/medicalHypernymy}}. All instances in our corpora are symptom and disease pairs of compound entities. The statistical characteristics of these two corpora are shown in Table \ref{table:Dataset}.

\begin{table}[!htbp]
\begin{center}
\renewcommand\arraystretch{1.1}
\caption{Statistical Characteristics of English and Chinese Corpora}
\label{table:Dataset}
\begin{tabular}{c|c|rrr}
\hline
\multicolumn{2}{c|}{corpus}           & \multicolumn{1}{c}{positive} & \multicolumn{1}{c}{negative} & \multicolumn{1}{c}{all}    \\
\hline
\multirow{3}{*}{English} & training set & 27,872        & 27,872        & 55,744      \\
                         & test set  & 9,954        & 9,954        & 19,908      \\
                         & validation set  &1,991        & 1,991        & 3,982      \\
\hline
\multirow{3}{*}{Chinese} & train set & 8,960    & 8,960    & 17,920 \\
                         & test set & 3,200    & 3,200    & 6,400  \\
                         & validation set & 640      & 640      & 1,280  \\
\hline
\end{tabular}
\end{center}
\end{table}

The English corpus is constructed by extracting clinical finding pairs in SNOMED CT~\cite{donnelly2006snomed}. Since there is no Chinese version of SNOMED CT, following Ruan et al.~\cite{Tong2017An}, we build a Chinese corpus by extracting hypernymy and synonymy relations between symptoms from semi-structured and unstructured data on the detail pages of six selected Chinese healthcare websites. We set hypernymy symptom pairs as positive instances, hyponymy, synonymy and unrelated symptom pairs as negative instances.

\subsection{Experimental settings}

We set $a = 1$ and $b = 0$ for our attention mechanism. Corresponding to the attention mechanism, each compound entity is required to ends with a special end-of-term symbol ``$\left \langle EOS \right \rangle$'', and the attention weight $w_i^{(k)}$ of ``$\left \langle EOS \right \rangle$'' is set to 1, which enables the model to always have a non-zero output vector of an entity.

Due to the lack of context corpus, we randomly initialize word embeddings and each embedding is 256-dimensional. Both the size of GRU hidden states and the dimension of the capsules are set to 64. We have routing between two consecutive capsule layers (i.e. entity capsules and classification capsules) with two iterations (i.e. $r = 2$), and all the routing logits are initialized to zero. The model is trained by an adaptive learning rate method AdaDelta~\cite{Zeiler2012ADADELTA} to minimize the margin loss and the batch size is 128.

In the following experiments, widely-used performance measures such as precision (P), recall (R), and F$_1$-score (F$_1$) \cite{Liu2014} are used to evaluate the methods.

\subsection{Comparison with state-of-the-art methods}

In the experiments, we compare our model with state-of-the-art methods. These reference algorithms can be roughly divided into three categories: two basic methods (i.e., string containing method and set containing method), three distributional methods (i.e., feature vector method~\cite{baroni2012entailment,roller2014inclusive}, projection learning method~\cite{fu2014learning} and simple RNN method~ \cite{jiang2017constructing}), and one distributed method (i.e., term embedding method~\cite{yu2015learning}).

\begin{table}[!ht]
\begin{center}
\renewcommand\arraystretch{1.1}
\begin{scriptsize}
\caption{Comparative Results of Our Model and Reference Methods}
\label{table:Comparison}
\begin{tabular}{c|c|ccc|ccc}
\hline
\multicolumn{2}{c|}{\multirow{2}{*}{methods}} & \multicolumn{3}{c|}{English corpus} & \multicolumn{3}{c}{Chinese corpus} \\ \cline{3-8}
\multicolumn{2}{c|}{} & P & R & F$_1$ & P & R & F$_1$ \\
\hline
\multicolumn{2}{c|}{string containing}& \textbf{95.20} & 2.39 & 4.66 & \textbf{99.17}  & 78.47 & 87.61\\
\multicolumn{2}{c|}{set containing}  & 94.45 & 5.64 & 10.66 & 97.95 & 86.66 & 91.96\\
\hline
\multirow{2}{*}{feature vector} & whole entity & 73.72 & 81.77 & 77.54 & 79.51 & 69.47 & 74.15 \\
& sum of words& 87.31 & 88.95 & 88.12 & 93.35 & \textbf{94.78}   & 94.06\\
\hline
projection & whole entity & 70.96 & 83.47 & 76.71 & 78.19 & 64.75 & 70.84\\
learning & sum of words& 83.30& 87.22& 85.21 & 85.32 & 86.84 & 86.08 \\
\hline
\multirow{2}{*}{simple RNN}& whole entity& 76.51 & 80.78 & 78.59 & 81.42 & 62.59 & 70.78 \\
& sum of words & 88.52 & 89.92 & 89.22 & 80.11 & 63.56  & 70.88  \\
\hline
term & whole entity & 29.81 & 59.62 & 39.75 & 78.19 & 64.75 & 70.84 \\
embedding & sum of words & 42.63 & 85.25 & 56.83  & 38.78 & 77.56 & 51.73 \\
\hline
\multicolumn{2}{c|}{our model} & 89.30 & \textbf{91.42} & \textbf{90.35} & 96.09    & 94.44 & \textbf{95.26} \\
\hline
\end{tabular}
\end{scriptsize}
\end{center}
\end{table}

Table \ref{table:Comparison} summarizes comparative results of our model and six reference algorithms on English and Chinese corpora. Note that as to some reference algorithms, we also have an adaptation that use the sum of word embeddings as the entity vector to solve the OOV problem. From the table, we clearly observe that our model outperforms these reference algorithms. More specifically, our attention-based Bi-GRU-CapsNet model achieves the best F$_1$-scores both on English and Chinese corpora compared with all reference algorithms. While for the performance in terms of recall and precision, our model is also highly competitive.

For two basic methods, we observe that they perform well on Chinese corpus, but get poor performance in terms of recall ($\leqslant 5.64\%$) and F$_1$-score ($\leqslant 10.66\%$) on English corpus. For the adapted reference algorithms, we observe that they achieve similar performance ($\geqslant 62.59\%$) in terms of all three measures except for term embedding method. Compared to the whole entity version, all four adapted reference algorithms with sum of words achieve better performance except term embedding method on Chinese corpus. Roughly speaking, our model achieves the better performance ($\geqslant 89.30\%$) than all these four adapted reference algorithms except for one exception. That is, the recall of feature vector method with sum of word embeddings on Chinese corpus is $94.78\%$, which is slightly better than the recall of our model (i.e., $94.44\%$). These interesting observations confirm the usefulness of our model for hypernymy detection between compound entities. Our model achieves highly competitive performance on both English and Chinese corpora compared with these six reference algorithms.

\subsection{Effectiveness of the attention mechanism}
\label{SubSec:Effectiveness of the Attention Mechanism}

\begin{table*}[!ht]
\begin{center}
\renewcommand\arraystretch{1.1}
\begin{scriptsize}
\caption{Comparative Results of Four Attention Mechanisms on English Corpus}
\label{table:Attention}
\begin{tabular}{c|c|c|c|c|c|c|c}
\hline
& $aw_i^{(k)}+b$ & softmax? & precision & recall & F$_1$-score & same parts & different parts\\
\hline
\multirow{3}{*}{unequal attention} & $a=1, b=0$ & no & \textbf{89.30} & \textbf{91.42} &      \textbf{90.35} & no attention & attention\\
 & $a=10, b=0$ & no & 88.88 & 91.35 & 90.10 & no attention &  much attention\\
 & $a=1, b=0$ & yes      & 85.74 & 90.00 & 87.82 & little attention & a little attention\\
\hline
\multirow{1}{*}{equal attention} & $a=0, b=1$ & no & 85.71 & 89.72 & 87.67 & equal attention & equal attention \\
\hline
\end{tabular}
\end{scriptsize}
\end{center}
\end{table*}

This section is devoted to investigating the effectiveness of our attention mechanism. Based on English corpus, we empirically compare following four attention mechanisms:
\begin{enumerate}
    \item Attention weight $\alpha_i^{(k)} = w_i^{(k)}$ without a softmax layer;
    \item Attention weight $\alpha_i^{(k)} = 10 w_i^{(k)}$ without a softmax layer;
    \item Attention weight $\alpha_i^{(k)} = w_i^{(k)}$ with a softmax layer;
    \item Attention weight $\alpha_i^{(k)} = 1$ without a softmax layer, i.e. equal attention.
\end{enumerate}

Table \ref{table:Attention} shows the comparative results of four different attention schemes on English corpus. The performance of our model with unequal attention is better than that of our model with equal attention (i.e. no special attention). We also observe that our model achieves the best performance when there is no attention used to the same parts between compound entities (i.e. $a = 1$ and $b = 0$ without softmax). In our model, we introduce an attention mechanism to focus on the differences between two compound entities. This strategy can effectively bring good generalization when an unseen pair have the different parts which have existed in the training set.

\section{Conclusion}
\label{Sec:Conclusion}

In this paper, we propose an attention-based Bi-GRU-CapsNet model for hypernymy detection between compound entities. Experimental results on both English and Chinese corpora of symptom and disease pairs show that our proposed model achieves significant improvements compared to existing methods in the literature. As future work, we plan to improve synonymy inference for hypernymy detection.

\section*{Acknowledgments}

We would like to thank the referees for their useful comments and suggestions. This work was supported by the National Natural Science Foundation of China (No. 61772201), the National Key R\&D Program of China for ``Precision Medical Research" (No.~2018YFC0910500) and National Major Scientific and Technological Special Project for ``Significant New Drugs Development" (No.~2018ZX09201008).

\bibliographystyle{IEEEtran}
\bibliography{mybibfiles}

\begin{thebibliography}{10}
\providecommand{\url}[1]{#1}
\csname url@samestyle\endcsname
\providecommand{\newblock}{\relax}
\providecommand{\bibinfo}[2]{#2}
\providecommand{\BIBentrySTDinterwordspacing}{\spaceskip=0pt\relax}
\providecommand{\BIBentryALTinterwordstretchfactor}{4}
\providecommand{\BIBentryALTinterwordspacing}{\spaceskip=\fontdimen2\font plus
\BIBentryALTinterwordstretchfactor\fontdimen3\font minus
  \fontdimen4\font\relax}
\providecommand{\BIBforeignlanguage}[2]{{%
\expandafter\ifx\csname l@#1\endcsname\relax
\typeout{** WARNING: IEEEtran.bst: No hyphenation pattern has been}%
\typeout{** loaded for the language `#1'. Using the pattern for}%
\typeout{** the default language instead.}%
\else
\language=\csname l@#1\endcsname
\fi
#2}}
\providecommand{\BIBdecl}{\relax}
\BIBdecl

\bibitem{Snow2006Semantic}
R.~Snow, D.~Jurafsky, and A.~Y. Ng, ``Semantic taxonomy induction from
  heterogenous evidence.'' in \emph{ACL 2006, Sydney, Australia, 17-21 July},
  2006, pp. 801--808.

\bibitem{McNamee2008}
P.~McNamee, R.~Snow, P.~Schone, and J.~Mayfield, ``Learning named entity
  hyponyms for question answering,'' in \emph{{IJCNLP} 2008, Hyderabad, India,
  January 7-12, 2008}, 2008, pp. 799--804.

\bibitem{Sogancioglu2017}
G.~Sogancioglu, H.~{\"{O}}zt{\"{u}}rk, and A.~{\"{O}}zg{\"{u}}r, ``{BIOSSES:} a
  semantic sentence similarity estimation system for the biomedical domain,''
  \emph{Bioinformatics}, vol.~33, no.~14, pp. i49--i58, 2017.

\bibitem{hearst1992automatic}
M.~A. Hearst, ``Automatic acquisition of hyponyms from large text corpora,'' in
  \emph{{COLING} 1992, Nantes, France, August 23-28, 1992}, 1992, pp. 539--545.

\bibitem{snow2005learning}
R.~Snow, D.~Jurafsky, and A.~Y. Ng, ``Learning syntactic patterns for automatic
  hypernym discovery,'' in \emph{Advances in Neural Information Processing
  Systems}, 2005, pp. 1297--1304.

\bibitem{shwartz2016improving}
V.~Shwartz, Y.~Goldberg, and I.~Dagan, ``Improving hypernymy detection with an
  integrated path-based and distributional method,'' in \emph{Proceedings of
  the 54th Annual Meeting of the Association for Computational Linguistics,
  {ACL} 2016, August 7-12, 2016, Berlin, Germany}, 2016, pp. 2389--2398.

\bibitem{kotlerman2010directional}
L.~Kotlerman, I.~Dagan, I.~Szpektor, and M.~Zhitomirsky-Geffet, ``Directional
  distributional similarity for lexical inference,'' \emph{Natural Language
  Engineering}, vol.~16, no.~4, pp. 359--389, 2010.

\bibitem{lenci2012identifying}
A.~Lenci and G.~Benotto, ``Identifying hypernyms in distributional semantic
  spaces,'' in \emph{Proceedings of the First Joint Conference on Lexical and
  Computational Semantics, *SEM 2012, June 7-8, 2012, Montr{\'{e}}al, Canada.},
  2012, pp. 75--79.

\bibitem{Fu2013Exploiting}
R.~Fu, B.~Qin, and T.~Liu, ``Exploiting multiple sources for open-domain
  hypernym discovery,'' in \emph{Proceedings of the 2013 Conference on
  Empirical Methods in Natural Language Processing, {EMNLP} 2013, 18-21 October
  2013, Grand Hyatt Seattle, Seattle, Washington, USA}, 2013, pp. 1224--1234.

\bibitem{yu2015learning}
Z.~Yu, H.~Wang, X.~Lin, and M.~Wang, ``Learning term embeddings for hypernymy
  identification,'' in \emph{Proceedings of the Twenty-Fourth International
  Joint Conference on Artificial Intelligence, {IJCAI} 2015, Buenos Aires,
  Argentina, July 25-31, 2015}, 2015, pp. 1390--1397.

\bibitem{Anh2016Learning}
L.~A. Tuan, Y.~Tay, S.~C. Hui, and S.~K. Ng, ``Learning term embeddings for
  taxonomic relation identification using dynamic weighting neural network,''
  in \emph{Conference on Empirical Methods in Natural Language Processing},
  2016, pp. 403--413.

\bibitem{wang2018using}
Q.~Wang, T.~Wang, and C.~Xu, ``Using a knowledge graph for hypernymy detection
  between chinese symptoms,'' in \emph{ICACI 2018}.\hskip 1em plus 0.5em minus
  0.4em\relax IEEE, 2018, pp. 601--606.

\bibitem{roller2014inclusive}
S.~Roller, K.~Erk, and G.~Boleda, ``Inclusive yet selective: Supervised
  distributional hypernymy detection,'' in \emph{{COLING} 2014, August 23-29,
  2014, Dublin, Ireland}, 2014, pp. 1025--1036.

\bibitem{fu2014learning}
R.~Fu, J.~Guo, B.~Qin, W.~Che, H.~Wang, and T.~Liu, ``Learning semantic
  hierarchies via word embeddings,'' in \emph{Proceedings of the 52nd Annual
  Meeting of the Association for Computational Linguistics, {ACL} 2014, June
  22-27, 2014, Baltimore, MD, USA}, 2014, pp. 1199--1209.

\bibitem{Glava2017Dual}
G.~Glavas and S.~P. Ponzetto, ``Dual tensor model for detecting asymmetric
  lexico-semantic relations,'' in \emph{{EMNLP} 2017, Copenhagen, Denmark,
  September 9-11, 2017}, 2017, pp. 1757--1767.

\bibitem{cho2014learning}
K.~Cho, B.~van Merrienboer, {\c{C}}.~G{\"{u}}l{\c{c}}ehre, F.~Bougares,
  H.~Schwenk, and Y.~Bengio, ``Learning phrase representations using {RNN}
  encoder-decoder for statistical machine translation,'' \emph{CoRR}, 2014.

\bibitem{Sabour2017Dynamic}
S.~Sabour, N.~Frosst, and G.~E. Hinton, ``Dynamic routing between capsules,''
  in \emph{Advances in Neural Information Processing Systems}, 2017, pp.
  3856--3866.

\bibitem{donnelly2006snomed}
K.~Donnelly, ``{SNOMED-CT}: The advanced terminology and coding system for
  ehealth.'' \emph{Studies in Health Technology \& Informatics}, vol. 121, no.
  121, p. 279, 2006.

\bibitem{Tong2017An}
T.~Ruan, M.~Wang, J.~Sun, T.~Wang, L.~Zeng, Y.~Yin, and J.~Gao, ``An automatic
  approach for constructing a knowledge base of symptoms in chinese,'' in
  \emph{{BIBM} 2016, {IEEE}, Shenzhen, China, December 15-18, 2016}, 2016, pp.
  1657--1662.

\bibitem{Zeiler2012ADADELTA}
M.~D. Zeiler, ``{ADADELTA:} an adaptive learning rate method,'' \emph{CoRR},
  vol. abs/1212.5701, 2012.

\bibitem{Liu2014}
Y.~Liu, Y.~Zhou, S.~Wen, and C.~Tang, ``A strategy on selecting performance
  metrics for classifier evaluation,'' \emph{{IJMCMC}}, vol.~6, no.~4, pp.
  20--35, 2014.

\bibitem{baroni2012entailment}
M.~Baroni, R.~Bernardi, N.~Q. Do, and C.~C. Shan, ``Entailment above the word
  level in distributional semantics,'' in \emph{{EACL} 2012, Avignon, France,
  April 23-27, 2012}, 2012, pp. 23--32.

\bibitem{jiang2017constructing}
T.~Jiang, M.~Liu, B.~Qin, and T.~Liu, ``Constructing semantic hierarchies via
  fusion learning architecture,'' in \emph{China Conference on Information
  Retrieval}.\hskip 1em plus 0.5em minus 0.4em\relax Springer, 2017, pp.
  136--148.

\end{thebibliography}

\end{CJK*}
\end{document}